\documentclass[10pt,letterpaper]{article}
\usepackage[top=0.85in,left=1.0in,footskip=0.75in]{geometry}

\usepackage{amsmath,amssymb}

\usepackage{changepage}

\usepackage[utf8x]{inputenc}

\usepackage{textcomp,marvosym}

\usepackage{cite}

\usepackage{nameref,hyperref}

\usepackage[right]{lineno}

\usepackage{microtype}
\DisableLigatures[f]{encoding = *, family = * }

\usepackage[table]{xcolor}

\usepackage{array}

\newcolumntype{+}{!{\vrule width 2pt}}

\newlength\savedwidth

\raggedright
\usepackage[aboveskip=1pt,labelfont=bf,labelsep=period,justification=raggedright,singlelinecheck=off]{caption}

\makeatletter
\renewcommand{\@biblabel}[1]{\quad#1.}
\makeatother

\date{}

\usepackage{lastpage,fancyhdr,graphicx}

\newcommand{\alfa}{$\mathit{Alpha}$}
\newcommand{\kalfa}{\mathit{Alpha}}
\newcommand{\rtavg}{\overline{\mathit{RT}}}

\usepackage{color}

\begin{document}
\vspace*{0.2in}

\begin{flushleft}
{\Large
\textbf\newline{Cohesion and Coalition Formation in the European Parliament:
Roll-Call Votes and Twitter Activities} 
}
\newline
\\
Darko Cherepnalkoski\textsuperscript{1}*,
Andreas Karpf\textsuperscript{2},
Igor Mozeti\v{c}\textsuperscript{1},
Miha Gr\v{c}ar\textsuperscript{1}*
\\
\bigskip
\textsuperscript{1} Department of Knowledge Technologies, Jo\v{z}ef Stefan Institute, Ljubljana, Slovenia
\\
\textsuperscript{2} Universit\'e Panth\'eon-Sorbonne / Paris School of Economics, Paris, France
\bigskip


* darko.cerepnalkoski@gmail.com (DC), miha.grcar@ijs.si (MG)
\end{flushleft}

\section*{Abstract}

We study the cohesion within and the coalitions between political groups in the 
Eighth European Parliament (2014--2019) by analyzing two entirely different aspects 
of the behavior of the Members of the European Parliament (MEPs) in 
the policy-making processes.
On one hand, we analyze their co-voting patterns and, on the other, 
their retweeting behavior. We make use of two diverse datasets in the analysis. 
The first one is the roll-call vote dataset, where cohesion is regarded as 
the tendency to co-vote within a group, and a coalition is formed when the 
members of several groups exhibit a high degree of co-voting agreement on
a subject. The second dataset comes from Twitter; it captures the retweeting 
(i.e., endorsing) behavior of the MEPs and implies cohesion 
(retweets within the same group)
and coalitions (retweets between groups) from a completely different perspective. 

We employ two different methodologies to analyze the cohesion and coalitions. 
The first one is based on Krippendorff's Alpha reliability, used to measure 
the agreement between raters in data-analysis scenarios, and the second one is based 
on Exponential Random Graph Models, often used in social-network analysis. 
We give general insights into the cohesion of political groups in 
the European Parliament, 
explore whether coalitions are formed in the same way for different policy areas, 
and examine to what degree the retweeting behavior of MEPs 
corresponds to their co-voting patterns. A novel and interesting aspect of our 
work is the relationship between the co-voting and retweeting patterns.


\section{Introduction}

Social-media activities often reflect phenomena that occur in other complex systems. 
By observing social networks and the content propagated through these networks, 
we can describe or even predict the interplay between the observed social-media 
activities and another complex system that is more difficult, if not impossible, 
to monitor. There are numerous studies reported in the literature that
successfully correlate social-media activities to phenomena like election 
outcomes \cite{Gayo2011elections,Eom2015volumedyn} or stock-price movements 
\cite{Bollen2011twitter,Ranco2015eventstudy}.

In this paper we study the cohesion and coalitions exhibited by political groups 
in the Eighth European Parliament (2014--2019). We analyze two entirely different 
aspects of how the Members of the European Parliament (MEPs) behave 
in policy-making processes. 
On one hand, we analyze their co-voting patterns and, on the other, 
their retweeting (i.e., endorsing) behavior.

We use two diverse datasets in the analysis: the roll-call votes and the Twitter data.
A roll-call vote (RCV) is a vote in the parliament in which the names of the MEPs 
are recorded along with their votes. The RCV data is available as part of the 
minutes of the parliament's plenary sessions. From this perspective, 
cohesion is seen as the tendency to co-vote (i.e., cast the same vote) 
within a group, and a coalition is formed when members of two or more groups 
exhibit a high degree of co-voting on a subject. 
The second dataset comes from Twitter. It captures the retweeting 
behavior of MEPs and implies cohesion (retweets within the same group) and 
coalitions (retweets between the groups) from a completely different perspective.

With over 300 million monthly active users and 500 million tweets posted daily, 
Twitter is one of the most popular social networks. Twitter allows its users to post 
short messages (tweets) and to follow other users. A user who follows another 
user is able to read his/her public tweets.
Twitter also supports other types of interaction, such as user mentions, 
replies, and retweets. Of these, retweeting is the most important activity 
as it is used to share and endorse content created by other users. 
When a user retweets a tweet, the information about the original author 
as well as the tweet's content are preserved, and the tweet is shared with 
the user's followers. Typically, users retweet content that they agree with 
and thus endorse the views expressed by the original tweeter. 

We apply two different methodologies to analyze the cohesion and coalitions. 
The first one is based on Krippendorff's \alfa\, \cite{Krippendorff2012} 
which  measures the agreement among observers, 
or voters in our case.
The second one is based on Exponential Random Graph Models (ERGM) \cite{ergm2008}. 
In contrast to the former, ERGM is a network-based approach and is often used 
in social-network analyses. 
Even though these two methodologies come with two different sets of 
techniques and are based on different assumptions, they provide consistent results.

The main contributions of this paper are as follows:

(i) We give general insights into the cohesion of political groups in the 
Eighth European Parliament, both overall and across different policy areas.

(ii) We explore whether coalitions are formed in the same way for 
different policy areas.

(iii) We explore to what degree the retweeting behavior of MEPs 
corresponds to their co-voting patterns. 

(iv) We employ two statistically sound methodologies and examine the extent 
to which the results are sensitive to the choice of methodology.
While the results are mostly consistent, we show that the difference
are due to the different treatment of non-attending and abstaining MEPs
by \alfa\, and ERGM.

\begin{figure}[!h]
\centering
\includegraphics[width=\textwidth]{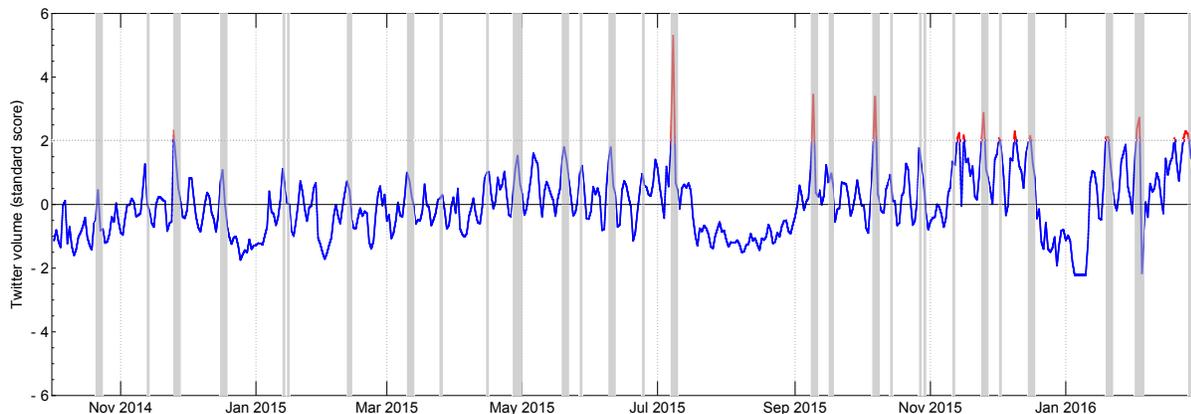}
\caption{{\bf Timeline of Twitter activity of MEPs and roll-call voting sessions.} 
Twitter activity is represented by the total number of tweets posted by 
all MEPs on a given day. The volume is standardized and the solid blue line 
represents the standard score. The peaks in Twitter activity that are above 
two standard deviations are highlighted in red. The shaded regions correspond 
to days with roll-call voting sessions.}
\label{fig:rcv_twitter_timeline}
\end{figure}

\begin{figure}[!h]
\centering
\includegraphics[width=\textwidth]{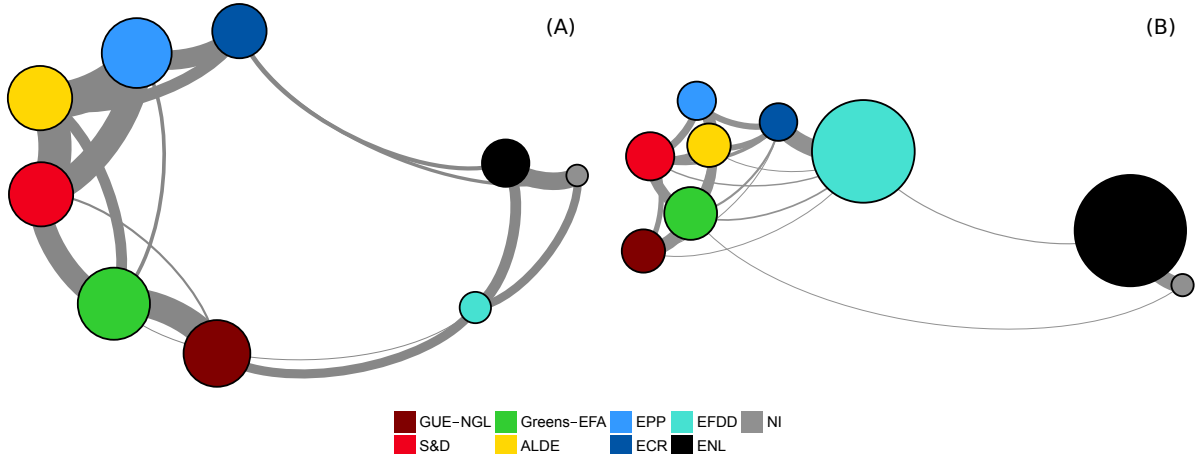}
\caption{{\bf Networks of roll-call votes and retweets.}
(A) Co-voting agreement within and between political groups. 
(B) Average retweets within and between political groups.}
\label{fig:covoting_retweeting_network}
\end{figure}

The most novel and interesting aspect of our work is the relationship between 
the co-voting and the retweeting patterns. The increased use of Twitter by MEPs 
on days with a roll-call vote session (see Fig~\ref{fig:rcv_twitter_timeline}) 
is an indicator that these two processes are related. In addition, the force-based 
layouts of the co-voting network and the retweet network reveal a very similar 
structure on the left-to-center side of the political spectrum 
(see Fig~\ref{fig:covoting_retweeting_network}). They also show a discrepancy 
on the far-right side of the spectrum, which calls for a more detailed analysis.

\subsection{Related work}

In this paper we study and relate two very different aspects of how MEPs behave 
in policy-making processes. First, we look at their co-voting behavior, and second, 
we examine their retweeting patterns. Thus, we draw related work from two 
different fields of science. On one hand, we look at how co-voting behavior is
analyzed in the political-science literature and, on the other, we explore 
how Twitter is used to better understand political and policy-making processes. 
The latter has been more thoroughly explored in the field of data mining 
(specifically, text mining and network analysis).

To the best of our knowledge, this is the first paper that studies legislative 
behavior in the Eighth European Parliament. The legislative behavior of the 
previous parliaments was thoroughly studied by Hix, Attina, and others
\cite{attina1990voting,kreppel1999coalition,quanjel1993growing,brzinski1995political,hix2002parliamentary,hix2009after}. 
These studies found that voting behavior is determined to a large extent---and 
when viewed over time, increasingly so---by affiliation to a political group, 
as an organizational reflection of the ideological position. The authors found that 
the cohesion of political groups in the parliament has increased, while nationality 
has been less and less of a decisive factor \cite{hix2007democratic}. 
The literature also reports that a split into political camps on the left and 
right of the political spectrum has recently replaced the 
`grand coalition' between the two big blocks of Christian Conservatives 
(\textit{EPP}) and Social Democrats (\textit{S\&D}) as the dominant form of 
finding majorities in the parliament. The authors conclude that coalitions 
are to a large extent formed along the left-to-right axis \cite{hix2007democratic}.

In this paper we analyze the roll-call vote data published in the minutes of 
the parliament's plenary sessions. For a given subject, the data contains 
the vote of each MEP present at the respective sitting.
Roll-call vote data from the European Parliament has already been 
extensively studied by other authors, most notably by 
Hix et al. \cite{hix2002parliamentary,hix2005power,hix2009after}. 
To be able to study the cohesion and coalitions, authors like Hix, Attina, 
and Rice \cite{attina1990voting,hix2005power,rice1928quantitative} defined 
and employed a variety of agreement measures. The most prominent measure is 
the Agreement Index proposed by Hix et al.\cite{hix2005power}. This measure 
computes the agreement score from the size of the majority class for a 
particular vote. The Agreement Index, however, exhibits two drawbacks: 
(i) it does not account for co-voting by chance, and (ii) without a 
proper adaptation, it does not accommodate the scenario in which 
the agreement is to be measured between two different political groups.

We employ two statistically sound methodologies developed in two different 
fields of science. The first one is based on Krippendorff's \alfa
\cite{Krippendorff2012}. \alfa\, is a measure of the agreement among observers, 
coders, or measuring instruments that assign values to items or phenomena. 
It compares the observed agreement to the agreement expected by chance. 
\alfa\, is used to measure the inter- and self-annotator agreement of human experts 
when labeling data, and the performance of classification models in machine learning 
scenarios \cite{Mozetic2016annot}. 
In addition to \alfa, we employ Exponential Random Graph Models (ERGM) \cite{ergm2008}.
In contrast to the former, ERGM is a network-based approach, often used in
social-network analyses. ERGM can be employed to investigate how 
different network statistics (e.g., number of edges and triangles)
or external factors (e.g., political group membership)
govern the network-formation process.

The second important aspect of our study is related to analyzing the behavior 
of participants in social networks, specifically Twitter. Twitter is studied 
by researchers to better understand different political processes, and in 
some cases to predict their outcomes. Eom et al. \cite{Eom2015volumedyn} 
consider the number of tweets by a party as a proxy for the collective attention 
to the party, explore the dynamics of the volume, and show that this quantity 
contains information about an election's outcome. 
Other studies \cite{Smailovic2015bulgelect}
reach similar conclusions. Conover et al. \cite{Conover2011alignment} predicted the 
political alignment of Twitter users in the run-up to the 2010 US elections 
based on content and network structure. They analyzed the polarization of the 
retweet and mention networks for the same elections \cite{Conover2011polarization}. 
Borondo et al. \cite{Borondo2012campaign} analyzed user activity during the 
Spanish presidential elections. They additionally analyzed the 2012 Catalan elections,
focusing on the interplay between the language and the community structure of the 
network \cite{Borondo2014patterns}. Most existing research, as Larsson points out
\cite{Larsson2014permcampaign}, focuses on the online behavior of leading political 
figures during election campaigns.

This paper continues our research on communities that MEPs (and their followers) 
form on Twitter \cite{Cherepnalkoski2016retweet}. The goal of our research was 
to evaluate the role of Twitter in identifying communities of influence when 
the actual communities are known. We represent the influence on Twitter by the 
number of retweets that MEPs ``receive''. We construct two networks of influence: 
(i) core, which consists only of MEPs, and (ii) extended, which also involves 
their followers. We compare the detected communities in both networks to the groups
formed by the political, country, and language membership of MEPs. 
The results show that the detected communities in the core network closely match 
the political groups, while the communities in the extended network correspond
to the countries of residence. This provides empirical evidence that analyzing 
retweet networks can reveal real-world relationships and can be used to 
uncover hidden properties of the networks.

Lazer\cite{Lazer2011} highlights the importance of network-based approaches
in political science in general by arguing that politics is a relational phenomenon 
at its core. Some researchers have adopted the network-based approach to investigate 
the structure of legislative work in the US Congress, including committee and 
sub-committee membership \cite{Porter2005netanalysis}, bill co-sponsoring
\cite{Zhang2008commstruct}, and roll-call votes \cite{Waugh2009polarization}. 
More recently, Dal Maso et al. \cite{DalMaso2014italparl} examined the community 
structure with respect to political coalitions and government structure in the 
Italian Parliament. Scherpereel et al. \cite{Scherpereel2016Adoption} examined the 
constituency, personal, and strategic characteristics of MEPs that influence 
their tweeting behavior. They suggested that Twitter's characteristics, like immediacy,
interactivity, spontaneity, personality, and informality, are likely to resonate with
political parties across Europe. By fitting regression models, the authors 
find that MEPs from incohesive groups have a greater tendency to retweet.

In contrast to most of these studies, we focus on the Eighth European Parliament, 
and more importantly, we study and relate two entirely different behavioral aspects,
co-voting and retweeting. The goal of this research is to better understand 
the cohesion and coalition formation processes in the European Parliament
by quantifying and comparing the co-voting patterns and social behavior.

\section{Data}

\paragraph*{Ethics statement.}
The tweets of MEPs were collected through the public
Twitter API and are subject to the Twitter terms and conditions.

\paragraph*{Data availability.}
All the data about roll-call votes and retweets of 
the Members of the Eighth European Parliament are available in a public 
language resource repository CLARIN.SI 
at \url{http://hdl.handle.net/11356/1071}.\\[2ex]

We collected the results of the roll-call votes (RCVs) from the website 
of the European Parliament \cite{eup2016rcv}. 
The RCV data is available in the form of XMLs.
From the same website, we also obtained the register of active 
MEPs \cite{eup2016mep}.
This register, even though it is available as XML data, presents a challenge when 
integrated with the RCV documents. The reason lies in the IDs of MEPs in 
the registry, which differ from the IDs of the same MEPs in the RCV documents. 
We used a name-matching technique based on Jaccard distance to establish the
missing mappings. Moreover, we linked each MEP to her/his Twitter ID and 
thus established an explicit link between the MEP and her/his tweeting activities.

The European Parliament consists of 751 members. Within the period of our study, 
from October 1, 2014 to February 29, 2016, 36 MEPs left the European Parliament and 
were substituted by 36 new MEPs. This gives a total of 787 MEPs that
participated in the work of the parliament during this period. 
In total, 79\% of the MEPs have Twitter accounts and 67\% have retweeted 
or been retweeted by another MEP. 
The distribution of MEPs by political groups, with their Twitter accounts,
and retweeting activity is given in Table~\ref{tab:groups_members}. 

\begin{table}[!h]
\caption{{\bf Distribution of Members of the European Parliament (MEPs) by 
political groups.}
There is the number of MEPs in each group, the number of MEPs with 
Twitter accounts, and the number of MEPs in the retweet network.
The numbers are cumulative for the period from October 1, 2014 
to February 29, 2016.
}
\label{tab:groups_members}
\centering
\begin{tabular}{l|r|rrrr}
Political &  Number     & Twitter  & Nodes in the \\
group     &  of MEPs    & accounts & retweet network \\
\hline
GUE-NGL     &  61 &  43 &  41 \\
S$\&$D      & 193 & 162 & 135 \\
Greens-EFA  &  51 &  48 &  44 \\
ALDE        &  77 &  62 &  54 \\
EPP         & 223 & 166 & 141 \\
ECR         &  82 &  64 &  48 \\
EFDD        &  45 &  39 &  34 \\
ENL         &  38 &  29 &  27 \\
NI          &  17 &   9 &   6 \\
\hline
Total       & 787 & 622 & 530 \\
\end{tabular}
\end{table}

Political groups are ordered by their seating positions in the European Parliament 
from the left to the right wing:
\begin{itemize}
\item \textit{GUE-NGL:} The European United Left–Nordic Green Left unite radical 
left wing, socialist and communist parties. They support a social union with stronger 
protection of workers and a higher degree of wealth redistribution.
\item\textit{S\&D:} The Socialists \& Democrats consist of social democratic, 
socialist and labour parties. They have a balanced
stance between workers protection and market economy, and are integration friendly.
\item \textit{Greens-EFA:} The Group of the Greens and European Free Alliance is the 
political home to the Green and regionalist/separatist political parties (EFA). 
The group is favorable to European integration, 
and strives for a more ecological and social union. 
\item \textit{ALDE:} The Alliance of Liberals and Democrats for Europe summons liberal
political groups. They are cooperation and integration friendly, with a stance 
against overregulation and bureaucratization.  
\item \textit{EPP:} The European People's Party 
unites christian democrats, conservatives and center-right parties. 
They are considered integration-friendly and strong supporters of the 
European Constitution.
\item \textit{ECR:} European Conservatives and Reformists provide a political home 
to conservative and Eurosceptic parties.
\item \textit{EFDD:} The Europe of Freedom and Direct Democracy 
unites extreme right-wing and Eurosceptic parties.
\item \textit{ENL:} The Europe of Nations and Freedom is the smallest 
group, with extreme right-wing members.
\item \textit{NI:} There are also some non-aligned (Non-Inscrits NI) 
MEPs, partly split-offs from national lists, or members of
other minority extreme-right wing political movements.
\end{itemize}

In the time frame of this study, the MEPs voted in 2535 roll-call votes. 
Of these, 2358 RCVs contain information about the policy area 
(referred to as \textit{subject} in the RCV metadata) they are part of. 
The distribution of RCVs by policy area is given in Table~\ref{tab:rcv_subjects}.

\begin{table}[!h]
\caption{{\bf Distribution of roll-call votes (RCVs) by policy areas.}
The time period is from October 1, 2014 to February 29, 2016.}
\label{tab:rcv_subjects}
\centering
\begin{tabular}{l|rrrrr}
Policy area & RCVs \\
\hline
Area of freedom, security and justice     &  116 \\
Community policies                        &  568 \\
Economic and monetary system              &  105 \\
Economic, social and territorial cohesion &  314 \\
European citizenship                      &   92 \\
External relations of the Union           &  512 \\
Internal market                           &  232 \\
State and evolution of the Union          &  419 \\
Unclassified                              &  177 \\
\hline
Total                                     & 2535 \\
\end{tabular}
\end{table}

From the data collected, a matrix of 787 MEPs by 2535 RCVs is formed.
Each entry is either \textit{yes} or \textit{no} (depending how the MEP voted
in the RCV), or missing (when the MEP abstained or did not attend the RCV session).
The matrix is used to compute the Krippendorff's \alfa\, co-voting agreement.
For the ERGM analysis, a network is formed for each RCV.
The nodes are all the 787 MEPs and
there is an edge between two MEPs if they cast the same vote (\textit{yes} or \textit{no}).
MEPs that abstained, or did not attend the RCV session,
are disconnected nodes.

The retweet network of the parliament consists of 530 nodes corresponding to the 
MEPs that are active on Twitter, and 4723 edges.
An undirected edge between two MEPs is formed when one of them
retweeted the other.
The weight of the edge
represents the number of times these MEPs have retweeted each other. 
The total weight of edges in the network, corresponding to the total number 
of retweets between the MEPs, is 26133.

\section{Methods}
\label{sec:methods}

In this section we present the methods to quantify cohesion and 
coalitions from the roll-call votes and Twitter activities.

\subsection{Co-voting measured by agreement}
\label{sec:methods_alpha}

We first show how the co-voting behaviour of MEPs can be quantified by
a measure of the agreement between them. We treat individual RCVs as observations,
and MEPs as independent observers or raters. When they cast the same vote,
there is a high level of agreement, and when they vote differently,
there is a high level of disagreement. We define cohesion as the level
of agreement within a political group, a coalition as a voting agreement 
between political groups, and opposition as a disagreement between different
groups.

There are many well-known measures of agreement in the literature.
We selected Krippendorff's Alpha-reliability (\alfa) \cite{Krippendorff2012},
which is a generalization of several specialized measures.
It works for any number of observers, and
is applicable to different variable types and metrics 
(e.g., nominal, ordered, interval, etc.).
In general, \alfa\, is defined as follows:
$$
\mathit{Alpha} = 1 - \frac{D_{o}}{D_{e}} \,,
$$
where $D_{o}$ is the actual disagreement between observers (MEPs), and
$D_{e}$ is disagreement expected by chance.
When observers agree perfectly, \alfa\,$=1$, when the
agreement equals the agreement by chance, \alfa\,$=0$, and
when the observers disagree systematically, \alfa\,$<0$.

The two disagreement measures are defined as follows:
$$
D_{o} = \frac{1}{N} \sum_{c,c'} N(c,c') \cdot \delta^2(c,c') \,,
$$
$$
D_{e} = \frac{1}{N(N-1)} \sum_{c,c'} N(c) \cdot N(c') \cdot \delta^2(c,c') \,.
$$
The arguments $N, N(c,c'), N(c)$, and $N(c')$ are defined below and
refer to the values in the coincidence matrix that is constructed
from the RCVs data. In roll-call votes, $c$ (and $c'$) is a nominal 
variable with two possible values: \textit{yes} and \textit{no}.
$\delta(c,c')$ is a difference function between the values of $c$ and $c'$,
defined as:
$$
\delta(c,c') =
  \begin{cases}
    0 & \text{ iff } c = c' \\
    1 & \text{ iff } c \neq c'
  \end{cases}
\;\;\;\; c,c' \in \{yes, no\} \;\;.
$$

The RCVs data has the form of a reliability data matrix:
$$
\begin{array}{rc|cccccc|}
\text{Roll-call votes} &  & 1      & 2      & \dots & u      & \dots & n \\
\hline
 & 1 & c_{11} & c_{12} & \dots & c_{1u} & \dots & c_{1n} \\
\text{Individual}  & . & .      & .      & \dots & .      & \dots & . \\
\text{MEPs voting} & i & c_{i1} & c_{i2} & \dots & c_{iu} & \dots & c_{in} \\
 & . & .      & .      & \dots & .      & \dots & . \\
 & m & c_{m1} & c_{m2} & \dots & c_{mu} & \dots & c_{mn} \\
\hline
\text{No. of MEPs voting} & & m_1    & m_2    & \dots & m_u    & \dots & m_n \\
\end{array}
$$
where $n$ is the number of RCVs, $m$ is the number of MEPs, $m_u$ is the number
of votes cast in the voting $u$, and $c_{iu}$ is the actual vote of an MEP $i$ in
voting $u$ (\textit{yes} or \textit{no}).

A coincidence matrix is constructed from the reliability data matrix,
and is in general a $k$-by-$k$ square matrix, where $k$ is the number of 
possible values of $c$.
In our case, where only \textit{yes/no} votes are relevant, the coincidence 
matrix is a $2$-by-$2$ matrix of the following form:
$$
\begin{array}{c|cc|c}
  &   & c' & \sum \\
\hline
  & . & . & . \\
c & . & N(c,c') & N(c) \\
\hline
\sum  & . & N(c') & N \\
\end{array}
$$

A cell $N(c,c')$ accounts for all coincidences from all pairs of MEPs in all 
RCVs where one MEP has voted $c$ and the other $c'$.
$N(c)$ and $N(c')$ are the totals for each vote outcome, and $N$ is the grand total.
The coincidences $N(c,c')$ are computed as:
$$
N(c,c') = \sum_u\frac{N_u(c,c')}{m_u-1}
\;\;\;\; c,c' \in \{yes, no\} 
$$
where $N_u(c,c')$ is the number of $(c,c')$ pairs 
in vote $u$, and $m_u$ is the number of MEPs that voted in $u$.
When computing $N_u(c,c')$, each pair of votes is considered twice,
once as a $(c,c')$ pair, and once as a $(c',c)$ pair.
The coincidence matrix is therefore symmetrical around the diagonal,
and the diagonal contains all the equal votes.

The \alfa\, agreement is used to measure the agreement between 
two MEPs or within a group of MEPs. When applied to a political group, 
\alfa\, corresponds to the cohesion of the group.
The closer \alfa\, is to 1, the higher the agreement of the MEPs in 
the group, and hence the higher the cohesion of the group.

We propose a modified version of \alfa\, to measure the agreement 
between two different groups, $A$ and $B$. In the case of a voting agreement
between political groups, high \alfa\, is interpreted as a coalition
between the groups, whereas negative \alfa\, indicates political opposition.

Suppose $A$ and $B$ are disjoint subsets of all the MEPs,
$A, B \subset \{1, \dots, m\}$,
$A \cap B = \{\}$.
The respective number of votes cast by both group members in vote $u$ 
is $m_u(A)$ and $m_u(B)$.
The coincidences are then computed as:
$$
N(c,c') = \sum_u\frac{N_u(c,c') \cdot (m_u(A) + m_u(B))}{2 \cdot m_u(A) \cdot m_u(B)}
\;\;\;\; c,c' \in \{yes, no\}
$$
where the $(c,c')$ pairs come from different groups, $A$ and $B$.
The total number of such pairs in vote $u$ is $m_u(A) \cdot m_u(B)$.
The actual number $N_u(c,c')$ of the pairs is multiplied by 
$\frac{m_u(A) + m_u(B)}{2 \cdot m_u(A) \cdot m_u(B)}$ so that the total contribution 
of vote $u$ to the coincidence matrix is $m_u(A) + m_u(B)$.

\subsection{A network-based measure of co-voting}
\label{sec:methods_ergm}

In this section we describe a network-based approach to analyzing the co-voting 
behavior of MEPs. For each roll-call vote we form a network, where
the nodes in the network are MEPs, and an undirected edge
between two MEPs is formed when they cast the same vote.

We are interested in the factors that determine the 
cohesion within political groups and coalition formation between political groups.
Furthermore, we investigate to what extent communication in a different social context,
i.e., the retweeting behavior of MEPs, can explain the co-voting of MEPs. 
For this purpose we apply an Exponential Random Graph Model \cite{ergm2008} to 
individual roll-call vote networks, and aggregate the results by means of the
meta-analysis.

\subsubsection{Exponential Random Graph Model (ERGM)}
\label{methodergm}
ERGMs allow us to investigate the factors relevant for the network-formation process. 
Network metrics, as described in the abundant literature, serve to gain information about 
the structural properties of the observed network. A model investigating the processes 
driving the network formation, however, has to take into account that there can be a multitude 
of alternative networks. If we are interested in the parameters influencing the network 
formation we have to consider all possible networks and measure their similarity to the 
originally observed  network. The family of ERGMs builds upon this idea.  

Assume a random graph $Y$, in the form of a binary adjacency matrix, made up of a set 
of $n$ nodes and $e$ edges $\{Y_{ij}: i=1,\dots,n; j=1,\dots,n\}$ where, similar to a binary 
choice model, $Y_{ij}=1$ if the nodes $(i,j)$ are connected and $Y_{ij}=0$ if not. Since  
network data is by definition relational and thus violates assumptions of independence, classical 
binary choice models, like logistic regression, cannot be applied in this context. 
Within an ERGM, the probability for a given network is modelled by 
\begin{align}
P(Y = y | \theta) = \frac{\exp(\theta^{T} \cdot g(y))}{c(\theta)} \;\;,
\label{eq:objfunc}
\end{align}
where $\theta$ is the vector of parameters and $g(y)$  
is the vector of network statistics (counts of network substructures), which are a function 
of the adjacency matrix $y$.
$c(\theta)=\sum \exp(\theta^T \cdot g(y))$ is a
normalization constant corresponding to the sample of all possible networks, 
which ensures a proper probability distribution. 
Evaluating the above expression allows us to make assertions if and how 
specific nodal attributes influence the network formation process. 
These nodal attributes can be endogenous (dyad-dependent parameters) to the network, 
like the in- and out-degrees of a node, or exogenous (dyad-independent parameters), 
as the party affiliation, or the country of origin in our case. 

\subsubsection{Interpretation of the coefficients}

An alternative formulation of the ERGM provides the interpretation of the coefficients.
We introduce the change statistic, which is defined as the change in the network statistics 
when an edge between nodes $i$ and $j$ is added or not. If $g(y^+_{ij})$ and $g(y^-_{ij})$ 
denote the vectors of counts of network substructures when the edge is added or not, 
the change statistics is defined as follows:
$$
\Delta y_{ij} = g(y^+_{ij}) - g(y^-_{ij}) \;\;.
$$
With this at hand it can be shown that the distribution of the 
variable $Y_{ij}$, conditional on the rest of the graph $Y_{ij}^c$, corresponds to:
$$
\log\left( \frac{p(Y_{ij}=1 | Y_{ij}^{c} = y_{ij}^{c})}{p(Y_{ij}=0 | Y_{ij}^{c} = y_{ij}^{c})}\right) = 
logit ( p(Y_{ij}=1 | Y_{ij}^{c} = y_{ij}^{c}) ) = \theta^T \Delta y_{ij} \;\;.
$$

This implies on the one hand that the probability depends on $y_{ij}^c$ via the 
change statistic $\Delta y_{ij}$, and on the other hand, that each coefficient 
within the vector $\theta$ represents an increase in the conditional log-odds 
($\mathit{logit}$) of the graph when the corresponding element in the vector $g(y)$ increases by one.
The need to condition the probability on the rest of the network can be illustrated 
by a simple example. The addition (removal) of a single edge alters the network 
statistics. If a network has only edges $(j,k)$ and $(i,k)$, the creation of 
an edge $(i,j)$ would not only add an additional edge but would also alter the 
count for other network substructures included in the model. In this example, 
the creation of the edge $(i,j)$ also increases the number of triangles by one.
The coefficients are transformed into probabilities with the logistic function:
$$
p = \frac{1}{1 + exp(-\theta)} \;\;.
$$
For example, in the context of roll-call votes, the probability
that an additional co-voting edge is formed between two nodes (MEPs)
of the same political group is computed with that equation. In this
context, the nodematch (nodemix) coefficients of the ERGM (described
in detail bellow) therefore refer to the degree of homophilous
(heterophilous) matching of MEPs with regard to their political
affiliation, or, expressed differently, the propensity of MEPs to
co-vote with other MEPs of their respective political group or another
group.

A positive coefficient reflects an increased chance that an edge
between two nodes with respective properties, like group affiliation,
given all other parameters unchanged, is formed. Or, put differently, a
positive coefficient implies that the probability of observing a
network with a higher number of corresponding pairs relative to the
hypothetical baseline network, is higher than to observe the baseline
network itself \cite{cranmer2011inferential}. For an intuitive
interpretation, log-odds value of $0$ corresponds to the even chance
probability of $0.5$. Log-odds of $+1$ correspond to an increase of
probability by $0.23$, whereas log-odds of $-1$ correspond to a
decrease of probability by $0.23$.

\subsubsection{Sampling and estimation}

The computational challenges of estimating ERGMs is to a large
degree due to the estimation of the normalizing constant. The number
of possible networks is already extremely large for very small
networks and the computation is simply not feasible. Therefore, an
appropriate sample has to be found, ideally covering the most probable
areas of the probability distribution. For this we make use of a
method from the Markov Chain Monte Carlo (MCMC) family, namely the
Metropolis-Hastings algorithm.

The idea behind this algorithm is to generate and sample highly
weighted random networks departing from the observed network. The
Metropolis-Hastings algorithm is an iterative algorithm which samples
from the space of possible networks by randomly adding or removing
edges from the starting network conditional on its density. If the
likelihood, in the ERGM context also denoted as weights, of the newly
generated network is higher than that of the departure network it is
retained, otherwise it is discarded. In the former case, the algorithm
starts anew from the newly generated network. Otherwise departure
network is used again. Repeating this procedure sufficiently often and
summing the weights associated to the stored (sampled) networks allows
to compute an approximation of the denominator in equation
\ref{eq:objfunc} (normalizing constant).

The algorithm starts sampling from the originally observed
network $G$. The optimization of the coefficients is done
simultaneously, equivalently with the Metropolis-Hastings algorithm. At the beginning
starting values have to be supplied. For the study at hand we used the
``ergm'' library from the statistical R software package
\cite{ergm2008} implementing the Gibbs-Sampling algorithm
\cite{robins2007} which is a special case of the Metropolis-Hastings
algorithm outlined.

\subsubsection{Specification of the ERGM}
In order to answer our question of the importance of the factors 
which drive the network formation process in the roll-call co-voting network, 
the ERGM is specified with the following parameters:

\begin{enumerate}
\item \textit{nodematch country:} This parameter adds one network
statistic to the model, i.e., the number of edges $(i,j)$ where
$country_i=country_j$. The coefficient indicates the homophilious
mixing behavior of MEPs with respect to their country of origin. In
other words, this coefficient indicates how relevant nationality
is in the formation of edges in the co-voting network.

\item \textit{nodematch national party:} This parameter adds one
network statistic to the model: the number of edges $(i,j)$ with
$national.party_i=national.party_j$. The coefficient indicates the
homophilious mixing behavior of the MEPs with regard to their party
affiliation at the national level. In the context of this study, this
coefficient can be interpreted as an indicator for within-party
cohesion at the national level.

\item \textit{nodemix EP group:} This parameter adds one network
statistic for each pair of European political groups.
These coefficients shed light on the degree of coalitions between different 
groups as well as the within group cohesion . Given that there are nine 
groups in the European Parliament, this coefficient adds in total 81 statistics
to the model. 

\item \textit{edge covariate Twitter:} This parameter corresponds to a
square matrix with the dimension of the adjacency matrix of the
network, which corresponds to the number of mutual retweets between
the MEPs.  It provides an insight about the extent to which
communication in one social context (Twitter), can explain cooperation
in another social context (co-voting in RCVs).
\end{enumerate}

\subsubsection{Meta-analysis}
An ERGM as specified above is estimated for each of the 2535
roll-call votes. Each roll-call vote is thereby interpreted as a
binary network and as an independent study. It is assumed that a
priori each MEP could possibly form an edge with each other MEP in the
context of a roll-call vote. Assumptions over the presence or absence
of individual MEPs in a voting session are not made. In other words
the dimensions of the adjacency matrix (the node set), and therefore
the distribution from which new networks are drawn, is kept constant
over all RCVs and therefore for every ERGM. The ERGM results therefore
implicitly generalize to the case where potentially all MEPs are
present and could be voting. Not voting is incorporated implicitly by
the disconnectedness of a node.

The coefficients of the 2535 roll-call vote studies are
aggregated by means of a meta-analysis approach proposed by Lubbers
\cite{lubbers2003group} and Snijders et al. \cite{snijders2003multilevel}.
We are interested in average effect sizes of different matching patterns
over different topics and overall. Considering the number of RCVs, it
seems straightforward to interpret the different RCV networks as
multiplex networks and collapse them into one weighted
network, which could then be analysed by means of a
valued ERGM \cite{krivitsky2012exponential}. There are, however, two
reasons why we chose the meta-analysis approach instead. First,
aggregating the RCV data results into an extremely dense network,
leading to severe convergence (degeneracy) problems for the
ERGM. Second, the RCV data contains information about the different
policy areas the individual votes were about. Since we are interested
in how the coalition formation in the European Parliament differs over
different areas, a method is needed that allows for an ex-post
analysis of the corresponding results. We therefore opted for 
the meta-analysis approach by Lubbers and Snijders et al.
This approach allows us to summarize the results by decomposing the
coefficients into average effects and (class)
subject-specific deviations. The different ERGM
runs for each RCV are thereby regarded as different studies with
identical samples that are combined to obtain a general overview of
effect sizes.  The meta-regression model is defined as:
$$
\theta_m = \mu_{\theta} + E_m \;\;.
$$
Here $\theta_m$ is a parameter estimate for class $m$, and
$\mu_{\theta}$ is the average coefficient. $E_m$ denotes the
normally distributed deviation of the class $m$ with a mean of
$0$ and a variance of $\sigma^2$. $E_m$ is the estimation error of the
parameter value $\theta_m$ from the ERGM. The meta-analysis model is
fitted by an iterated, weighted, least-squares model in which the
observations are weighted by the inverse of their variances.
For the overall nodematch between political groups, we weighted the
coefficients by group sizes. The results from the meta analysis can be
interpreted as if they stemmed from an individual ERGM run.

In our study, the meta-analysis was performed 
using the RSiena library \cite{ripley2011manual}, which implements the method
proposed by Lubbers and Snijders et al. \cite{lubbers2003group,snijders2003multilevel}.

\subsection{Measuring cohesion and coalitions on Twitter}
\label{sec:methods_twitter}

The retweeting behavior of MEPs is captured by their retweet network. 
Each MEP active on Twitter is a node in this network. An edge in the network 
between two MEPs exists when one MEP retweeted the other. 
The weight of the edge is the number of retweets between the two MEPs. 
The resulting retweet network is an undirected, weighted network.

We measure the cohesion of a political group $A$ as the average retweets, 
i.e., the ratio of the number of retweets between the MEPs in 
the group $\mathit{retweets}(A)$ to the number of MEPs in the group $|A|$. 
The higher the ratio, the more each MEP (on average) retweets the MEPs from 
the same political group, hence, the higher the cohesion of the political group. 
The definition of the \textit{average retweeets} ($\rtavg$) of a group $A$ is:
$$
\rtavg(A)=\frac{\mathit{retweets}(A)}{|A|} \;\;.
$$
This measure of cohesion captures the aggregate retweeting behavior of the group. 
If we consider retweets as endorsements, a larger number of retweets within 
the group is an indicator of agreement between the MEPs in the group. 
It does not take into account the patterns of retweeting within the group, 
thus ignoring the social sub-structure of the group. This is a potentially 
interesting direction and we leave it for future work.

We employ an analogous measure for the strength of coalitions in the retweet network. 
The coalition strength between two groups $A$ and $B$ is the ratio 
of the number of retweets from one group to the other (but not within groups)
$\mathit{retweets}(A,B)$ to the total number of MEPs in both groups, $|A|+|B|$. 
The definition of the \textit{average retweeets} ($\rtavg$) between groups 
$A$ and $B$ is:
$$
\rtavg(A,B)=\frac{\mathit{retweets}(A,B)}{|A|+|B|} \;\;.
$$

\section{Results}
\label{sec:results}

\subsection{Cohesion of political groups}
\label{partystatecohesion}

In this section we first report on the level of cohesion of the European Parliament's 
groups by analyzing the co-voting through the agreement and ERGM measures. 
Next,  we explore two important policy areas, namely 
\textit{Economic and monetary system} 
and \textit{State and evolution of the Union}. 
Finally, we analyze the cohesion of the European Parliament's groups on Twitter.

\subsubsection{Co-voting cohesion}

Existing research by Hix et al. \cite{hix2002parliamentary,hix2005power,hix2009after} 
shows that the cohesion of the European political groups has been rising since the 1990s, 
and the level of cohesion remained high even after the EU's enlargement in 2004, 
when the number of MEPs increased from 626 to 732.

\begin{figure}[!h]
\centering
\includegraphics[width=\textwidth]{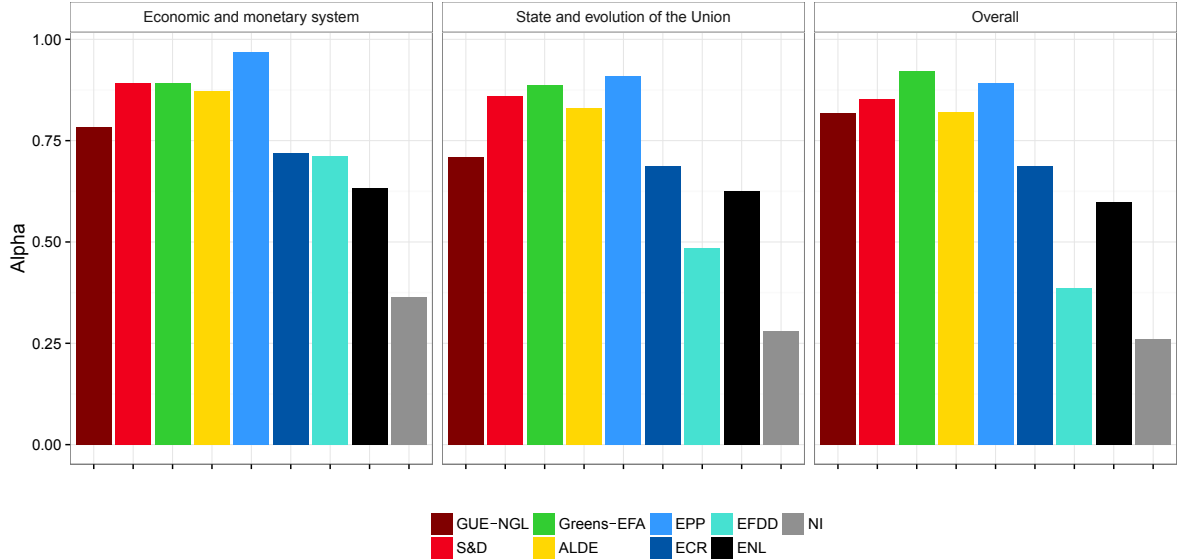}
\caption{{\bf Cohesion of political groups in terms of RCVs as measured by 
Krippendorff's \alfa.}
There are two selected policy areas 
(the left-hand panels), and overall cohesion across all policy areas 
(the right-hand panel).
The \alfa-agreement of $1$ indicates perfect co-voting agreement,
and $0$ indicates co-voting by chance.
The overall average \alfa\, across all nine political groups is $0.7$.
}
\label{fig:cohesion_party_alpha}
\end{figure}

We measure the co-voting cohesion of the political
groups in the Eighth European Parliament using 
Krippendorff's \textit{Alpha}---the results are shown 
in Fig~\ref{fig:cohesion_party_alpha} (panel \textit{Overall}). 
The \textit{Greens-EFA} have the highest cohesion of all the groups. 
This finding is in line with an analysis of previous compositions of the Fifth 
and Sixth European Parliaments
 by Hix and Noury \cite{hix2009after}, and the Seventh by 
VoteWatch \cite{Votewatch2011seventh}.
They are closely followed by the \textit{S\&D} and \textit{EPP}. 
Hix and Noury reported on the high cohesion of \textit{S\&D} in the Fifth 
and Sixth European Parliaments, 
and we also observe this in the current composition.
They also reported a slightly less cohesive \textit{EPP-ED}. 
This group split in 2009 into \textit{EPP} and \textit{ECR}. 
VoteWatch reports \textit{EPP} to have cohesion on a par with \textit{Greens-EFA} 
and \textit{S\&D} in the Seventh European Parliament. The cohesion level 
we observe in the current European Parliament 
is also similar to the level of \textit{Greens-EFA} and \textit{S\&D}.

The catch-all group of the non-aligned (\textit{NI}) comes out as the group with 
the lowest cohesion. In addition, among the least cohesive groups in 
the European Parliament 
are the Eurosceptics \textit{EFDD}, which include the British \textit{UKIP} 
led by Nigel Farage, and the \textit{ENL} whose largest party are the French 
\textit{National Front}, led by Marine Le Pen. 
Similarly, Hix and Noury found that the least cohesive groups in the 
Seventh European Parliament are the 
nationalists and Eurosceptics. The Eurosceptic \textit{IND/DEM}, which participated 
in the Sixth European Parliament, transformed into the current \textit{EFDD}, 
while the nationalistic \textit{UEN} was dissolved in 2009.

\begin{figure}[!h]
\centering
\includegraphics[width=\textwidth]{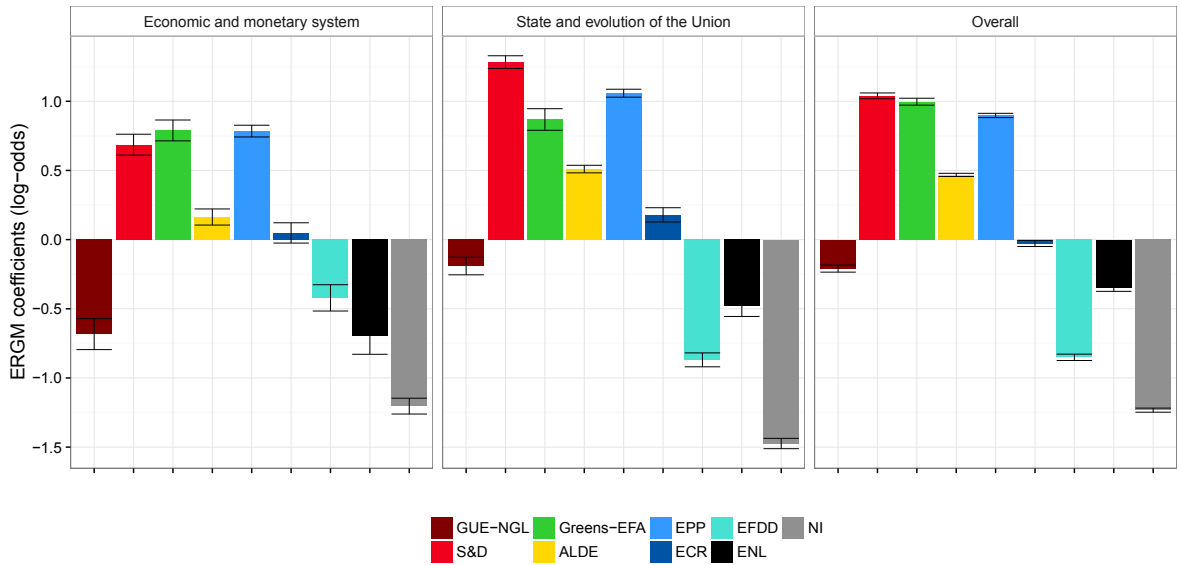}
\caption{{\bf Cohesion of political groups in terms of RCVs as measured by ERGM.}
There are two selected policy areas (the left-hand panels), and overall 
cohesion across all policy areas (the right-hand panel).
The coefficients of $0$ indicate baseline, i.e., an even chance of the same
votes within a group. Log-odds of $+1$ ($-1$) correspond to an increase
(decrease) of probability for $0.23$ of co-voting together.
}
\label{fig:cohesion_party_ergm}
\end{figure}

We also measure the voting cohesion of the European Parliament groups using an 
ERGM, a network-based 
method---the results are shown in Fig~\ref{fig:cohesion_party_ergm} 
(panel \textit{Overall}).
The cohesion results obtained with 
ERGM are comparable to the results based on agreement.
In this context, the parameters estimated by the ERGM refer to the matching of MEPs 
who belong to the same political group (one parameter per group). 
The parameters measure the homophilous matching 
between MEPs who have the same political affiliation. 
A positive value for the estimated parameter indicates that the co-voting of MEPs 
from that group is greater than what is expected by chance, where the expected number 
of co-voting links by chance in a group is taken to be uniformly random. 
A negative value indicates that there are fewer co-voting links within a group 
than expected by chance.

Even though \alfa\, and ERGM compute scores relative to what is expected by chance, 
they refer to different interpretations of chance. \alfa's concept of chance 
is based on the number of expected pair-wise co-votes between MEPs belonging 
to a group, knowing the votes of these MEPs on all RCVs. ERGM's concept 
of chance is based on the number of expected pair-wise co-votes between 
MEPs belonging to a group on a given RCV, knowing the network-related 
properties of the co-voting network on that particular RCV.
The main difference between \alfa\, and ERGM, though, is the treatment
of non-voting and abstained MEPs. \alfa\, considers only the yes/no votes,
and consequently, agreements by the voting MEPs of the same groups are
considerably higher than co-voting by chance.
ERGM, on the other hand, always considers all MEPs, and non-voting and abstained
MEPs are treated as disconnected nodes. The level of co-voting by chance
is therefore considerably lower, since there is often a large fraction of
MEPs that do not attand or abstain.

As with \alfa,
\textit{Greens-EFA}, \textit{S\&D}, and \textit{EPP} exhibit the 
highest cohesion, even though their ranking is permuted when compared to the 
ranking obtained with \alfa. At the other end of the scale, we observe the 
same situation as with \alfa. The non-aligned members \textit{NI} have 
the lowest cohesion, followed by \textit{EFDD} and \textit{ENL}.

The only place where the two methods disagree is the level of cohesion 
of \textit{GUE-NGL}. The \textit{Alpha} attributes \textit{GUE-NGL} a 
rather high level of cohesion, on a par with \textit{ALDE}, whereas the 
ERGM attributes them a much lower cohesion. 
The reason for this difference is the relatively high abstention
rate of \textit{GUE-NGL}. Whereas the overall fraction of non-attending
and abstaining MEPs across all RCVs and all political groups is 25\%,
the \textit{GUE-NGL} abstention rate is 34\%. This is reflected in an
above average cohesion by \alfa\, where only yes/no votes are considered,
and in a relatively lower, below average cohesion by ERGM.
In the later case, the non-attendance is interpreted as a non-cohesive
voting of a political groups as a whole.

In addition to the overall cohesion, we also focus on two selected policy areas.
The cohesion of the political groups related to these two policy areas is shown in
the first two panels in Fig~\ref{fig:cohesion_party_alpha} (\alfa) and
Fig~\ref{fig:cohesion_party_ergm} (ERGM).

The most important observation is that the level of cohesion of the 
political groups is very stable across different policy areas. 
These results are corroborated by both methodologies. 
Similar to the overall cohesion, the most cohesive political groups 
are the \textit{S\&D}, \textit{Greens-EFA}, and \textit{EPP}. 
The least cohesive group is the \textit{NI}, followed by 
the \textit{ENL} and \textit{EFDD}. The two methodologies agree 
on the level of cohesion for all the political groups, except for \textit{GUE-NGL},
due to a lower attendance rate.

\subsubsection{Cohesion on Twitter}

\begin{figure}[!h]
\centering
\includegraphics[width=12cm]{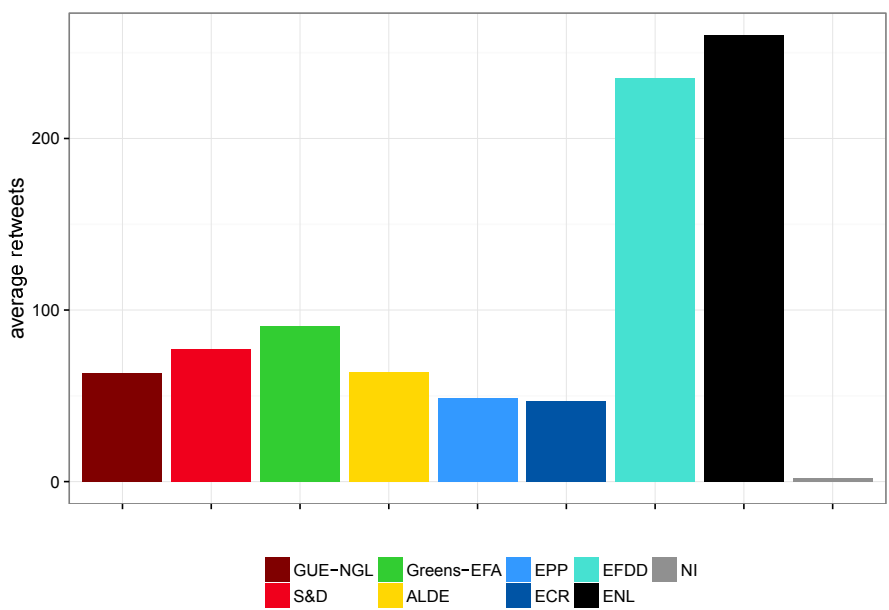}
\caption{{\bf Cohesion of political groups as estimated by retweeting within groups.}
The average number of retweets within the nine groups is $99$.
}
\label{fig:cohesion_twitter}
\end{figure}

We determine the cohesion of political groups on Twitter by using the average 
number of retweets between MEPs within the same group.
The results are shown in Fig~\ref{fig:cohesion_twitter}. 
The right-wing \textit{ENL} and \textit{EFDD} come out as the most cohesive groups,
while all the other groups have a far lower average number of retweets.
MEPs from \textit{ENL} and \textit{EFDD} post 
by far the largest number of retweets (over 240), and at the same time 
over 94\% of their retweets are directed to MEPs from the same group. 
Moreover, these two groups stand out in the way the retweets are distributed 
within the group. 
A large portion of the retweets of \textit{EFDD} (1755) go to Nigel Farage, 
the leader of the group. Likewise, a very large portion of retweets 
of \textit{ENL} (2324) 
go to Marine Le Pen, the leader of the group. Farage and Le Pen are by far the 
two most retweeted MEPs, with the third one having 
only 666 retweets.

\subsection{Coalitions in the European Parliament}

Coalition formation in the European Parliament is largely determined by ideological positions, 
reflected in the degree of cooperation of parties at the national and European levels. 
The observation of ideological inclinations in the coalition formation within 
the European Parliament
was already made by other authors \cite{hix2009after} and is confirmed in this study.
The basic patterns of coalition formation in the European Parliament 
can already be seen in the 
co-voting network in Fig~\ref{fig:covoting_retweeting_network}A.
It is remarkable that the degree of attachment between the 
political groups, 
which indicates the degree of cooperation in the European Parliament, 
nearly exactly corresponds to the 
left-to-right seating order. 

The liberal \textit{ALDE} seems to have an intermediator role between the left and 
right parts of the spectrum in the parliament. 
Between the extreme (\textit{GUE-NGL}) and center left (\textit{S\&D}) 
groups, this function seems to be occupied by 
\textit{Greens-EFA}. 
The non-aligned members \textit{NI},
as well as the Eurosceptic \textit{EFFD} and \textit{ENL}, 
seem to alternately tip the balance 
on both poles of the political spectrum. Being ideologically more inclined to vote 
with other conservative and right-wing groups (\textit{EPP}, \textit{ECR}), 
they sometimes also cooperate 
with the extreme left-wing group (\textit{GUE-NGL}) with which they share their 
Euroscepticism as a common denominator.

\subsubsection{Co-voting coalitions}

\begin{figure}[!h]
\centering
\includegraphics[width=\textwidth]{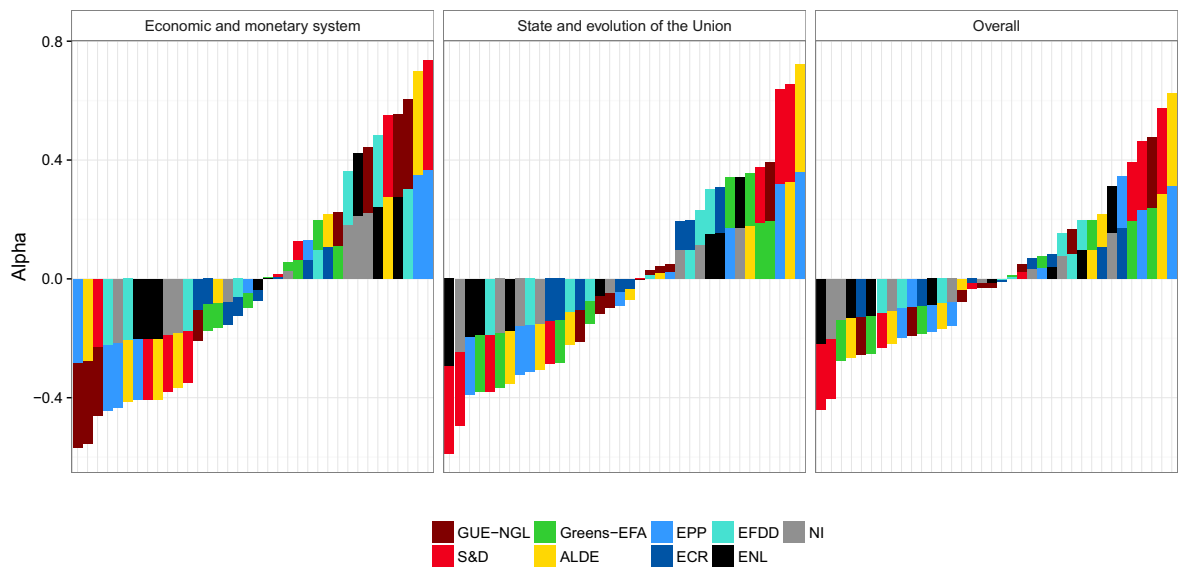}
\caption{{\bf Coalitions between political groups in terms of RCVs as measured by
Krippendorff's \alfa.}
For nine groups, there are 36 pairs for which we measure their (dis)agreement.
Positive values of \alfa\, indicate co-voting agreement,
negative values correspond to systematic disagreement, while
$\kalfa=0$ indicates co-voting by chance.
The average \alfa\, across all 36 pairs is $0.02$, close to co-voting by chance.
Note, however, that \alfa\, considers just yes/no votes, and ignores abstentions.
}
\label{fig:coalition_overall_ka}
\end{figure}

\begin{figure}[!h]
\centering
\includegraphics[width=\textwidth]{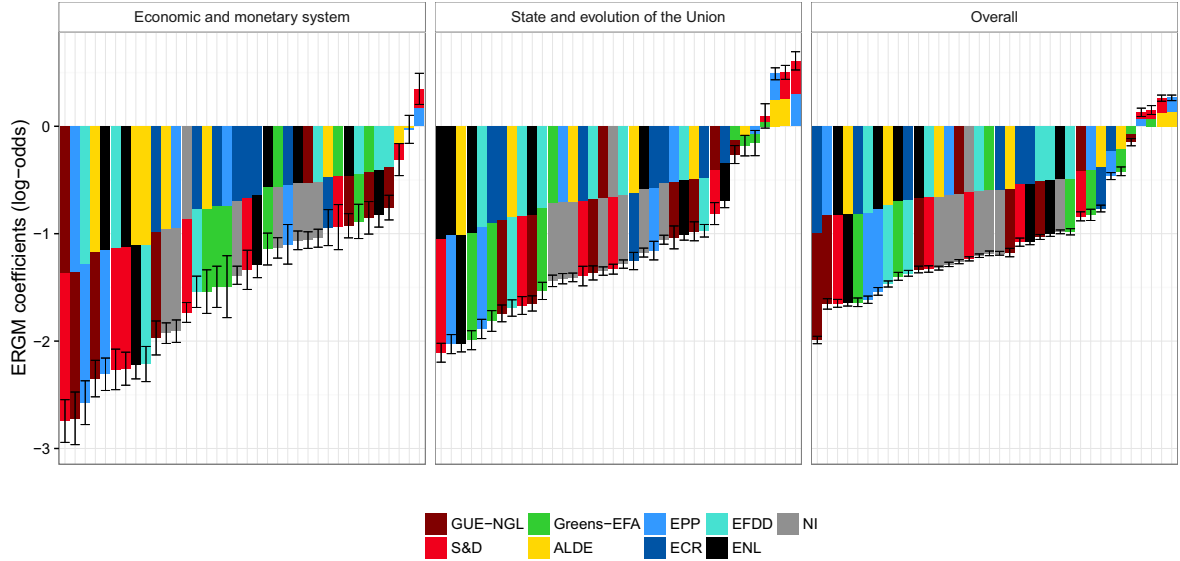}
\caption{{\bf Coalitions between political groups in terms of RCVs as measured by ERGM.}
There are 36 possible pairwise coalitions of the nine groups.
The coefficients of 0 indicate baseline, i.e., an even chance ($0.5$) of the two
groups to co-vote together. For most group pairs, the probability to co-vote is
lower than the chance: log-odds of $-1$ correspond to the probability of $0.27$,
log-odds of $-2$ to the probability of $0.12$, and
log-odds of $-3$ to the probability of $0.05$.
Note that ERGM does take abstentions into account, and therefore the baseline
of co-voting by chance is considerably higher than for \alfa.
}
\label{fig:coalition_overall_ergm}
\end{figure}

Figs \ref{fig:coalition_overall_ka} and \ref{fig:coalition_overall_ergm} give
a more detailed understanding of the coalition formation in the European Parliament. 
Fig \ref{fig:coalition_overall_ka} displays the degree of agreement or 
cooperation between political groups measured by Krippendorff's \alfa, whereas 
Fig \ref{fig:coalition_overall_ergm} is based on the result from the ERGM.
We first focus on the overall results displayed in the right-hand plots of 
Figs \ref{fig:coalition_overall_ka} and \ref{fig:coalition_overall_ergm}. 

The strongest degrees of cooperation 
are observed, with both methods, between the two major parties (\textit{EPP} and \textit{S\&D}) 
on the one hand, and the liberal \textit{ALDE} on the other. 
Furthermore, we see a strong propensity for 
\textit{Greens-EFA} to vote with the 
Social Democrats (5th strongest coalition by \alfa, and 3rd by ERGM) 
and the 
\textit{GUE-NGL} (3rd strongest coalition by \alfa, and 5th by ERGM).
These results underline the role of \textit{ALDE} and \textit{Greens-EFA} 
as intermediaries for the 
larger groups to achieve a majority. Although the two largest groups together 
have 405 seats and thus significantly more than the 376 votes 
needed for a simple majority, 
the degree of cooperation between the two major groups is ranked only 
as the fourth strongest 
by both methods. This suggests that these two political 
groups find it easier to negotiate deals with smaller counterparts than 
with the other large group. This observation was also made by 
Hix et al. \cite{hix2007democratic}, 
who noted that alignments on the left and right of the political spectrum 
have in recent years replaced the ``Grand Coalition'' between the two 
large blocks of Christian Conservatives (\textit{EPP}) and 
Social Democrats (\textit{S\&D}) 
as the dominant form of finding majorities in the parliament.

\subsubsection{Coalitions within individual policy areas}
\label{sec:coalitionpolicy}

Next, we focus on the coalition formation within the two selected policy areas.
The area \textit{State and Evolution of the Union} is
dominated by cooperation between the two major groups, \textit{S\&D} and \textit{EPP}, 
as well as \textit{ALDE}. 
We also observe a high degree of cooperation between groups that are generally 
regarded as integration friendly, like \textit{Greens-EFA} and \textit{GUE-NGL}. 
We see, particularly in Fig~\ref{fig:coalition_overall_ka}, 
a relatively high degree of cooperation between groups considered as Eurosceptic,
like \textit{ECR}, \textit{EFFD}, \textit{ENL}, and the group of non-aligned members. 

The dichotomy between supporters and opponents of European integration is even 
more pronounced within the policy area \textit{Economic and Monetary System}.
In fact, we are taking a closer look specifically at these two areas as they 
are, at the same time, both contentious and important.
Both methods rank the cooperation between 
\textit{S\&D} and \textit{EPP} on the one hand, and \textit{ALDE} on the other, 
as the strongest.

We also observe a certain degree of unanimity among the Eurosceptic and right-wing
groups (\textit{EFDD}, \textit{ENL}, and \textit{NI}) in this policy area. 
This seems plausible, 
as these groups were (especially in the aftermath of the global financial crisis 
and the subsequent European debt crisis) in fierce opposition to further payments
to financially troubled member states. However, we also observe a number 
of strong coalitions that might, at first glance, seem unusual, 
specifically involving the left-wing group \textit{GUE-NGL} on the one hand, and 
the right-wing \textit{EFDD}, \textit{ENL}, and \textit{NI} on the other. 
These links also show up in the network plot 
in Fig~\ref{fig:covoting_retweeting_network}A.
This might be attributable to a certain degree of Euroscepticism on both sides:
rooted in criticism of capitalism on the left, and at least partly a 
\textit{raison d'\^etre} on the right. 
Hix et al. \cite[p. 168]{hix2009after} discovered this pattern as well, 
and proposed an additional explanation---these coalitions also relate to a form 
of government-opposition dynamic that is rooted at the national level, 
but is reflected in voting patterns at the European level.

In general, we observe two main differences between the \alfa\, and
ERGM results: the baseline cooperation as estimated by \alfa\, is higher,
and the ordering of coalitions from the strongest to the weakest is not exactly
the same. The reason is the same as for the cohesion, namely different
treatment of non-voting and abstaining MEPs. When they are ignored,
as by \alfa, the baseline level of inter-group co-voting is higher.
When non-attending and abstaining is treated as voting differently,
as by ERGM, it is considerably more difficult to achieve co-voting coalitions, 
specially when there are on average 25\% MEPs that do not attend or abstain.
Groups with higher non-attendance rates, such as \textit{GUE-NGL} (34\%)
and \textit{NI} (40\%) are less likely to form coalitions, and therefore
have relatively lower ERGM coefficients (Fig \ref{fig:coalition_overall_ergm}) 
than \alfa\, scores (Fig \ref{fig:coalition_overall_ka}).

\subsubsection{Coalitions on Twitter}

\begin{figure}[!h]
\centering
\includegraphics[width=12cm]{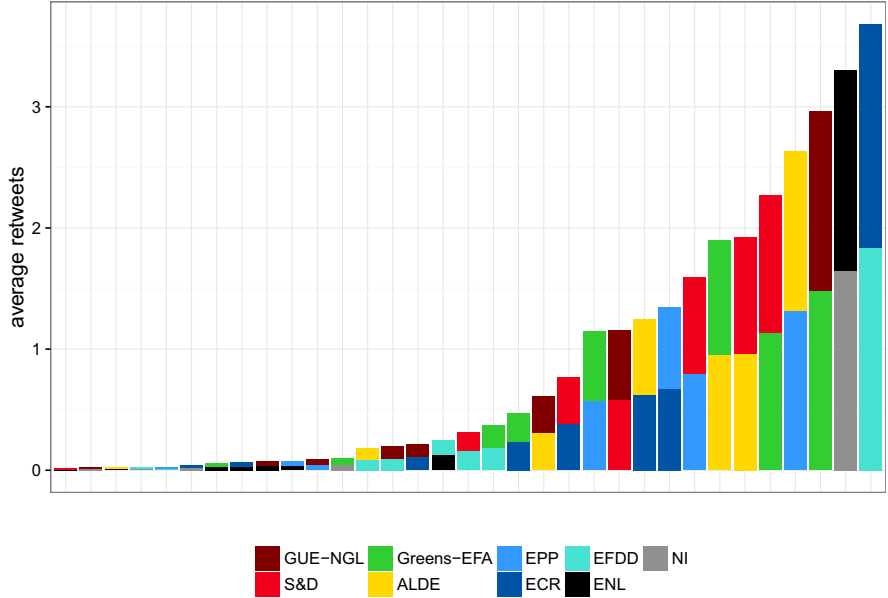}
\caption{{\bf Coalitions between political groups as estimated by retweeting 
between members of different groups.}
The average number of retweets between different 
political groups is $0.8$.
}
\label{fig:twitter_inter}
\end{figure}

The first insight into coalition formation on Twitter can be observed in the 
retweet network in Fig~\ref{fig:covoting_retweeting_network}B. 
The ideological left to right alignment of the political groups is reflected 
in the retweet network. Fig~\ref{fig:twitter_inter} shows the strength of 
the coalitions on Twitter, as estimated by the number of retweets between MEPs 
from different groups. 
The strongest coalitions are formed between the right-wing groups \textit{EFDD} 
and \textit{ECR}, as well as \textit{ENL} and \textit{NI}. At first, this might 
come as a surprise, since these groups do not form strong coalitions in the European Parliament,
as can be seen in Figs~\ref{fig:coalition_overall_ka} and \ref{fig:coalition_overall_ergm}.
On the other hand, the MEPs from these groups are very active Twitter users. 
As previously stated, MEPs from \textit{ENL} and \textit{EFDD} post the largest 
number of retweets. Moreover, 63\% of the retweets outside of \textit{ENL} 
are retweets of \textit{NI}. This effect is even more pronounced with MEPs from
\textit{EFDD}, whose retweets of \textit{ECR} account for 74\% of their
retweets from other groups.

In addition to these strong coalitions on the right wing, we find coalition 
patterns to be very similar to the voting coalitions observed in the European Parliament, seen in
Figs~\ref{fig:coalition_overall_ka} and \ref{fig:coalition_overall_ergm}. 
The strongest coalitions, which come immediately after the right-wing coalitions, 
are between \textit{Greens-EFA} on the one hand, and \textit{GUE-NGL} and \textit{S\&D} 
on the other, as well as \textit{ALDE} on the one hand, and \textit{EPP} and \textit{S\&D} 
on the other. These results corroborate the role of \textit{ALDE} and 
\textit{Greens-EFA} as intermediaries in the European Parliament, 
not only in the legislative process, 
but also in the debate on social media.

\subsubsection{Formation of coalitions}

To better understand the formation of coalitions in the European Parliament and on Twitter, 
we examine the strongest cooperation between political groups at three
different thresholds. For co-voting coalitions in the European Parliament 
we choose a high threshold 
of $\kalfa > 0.75$, a medium threshold of $\kalfa > 0.25$, and a negative 
threshold of $\kalfa < -0.25$ (which corresponds to strong oppositions). 
In this way we observe the overall patterns of coalition and opposition formation 
 in the European Parliament and in the two specific policy areas.
For cooperation on Twitter, we choose a high threshold of $\rtavg > 10$, 
a medium threshold of $\rtavg > 1$, and a very low threshold of $\rtavg < 0.001$.

\begin{figure}[!h]
\centering
\includegraphics[width=\textwidth]{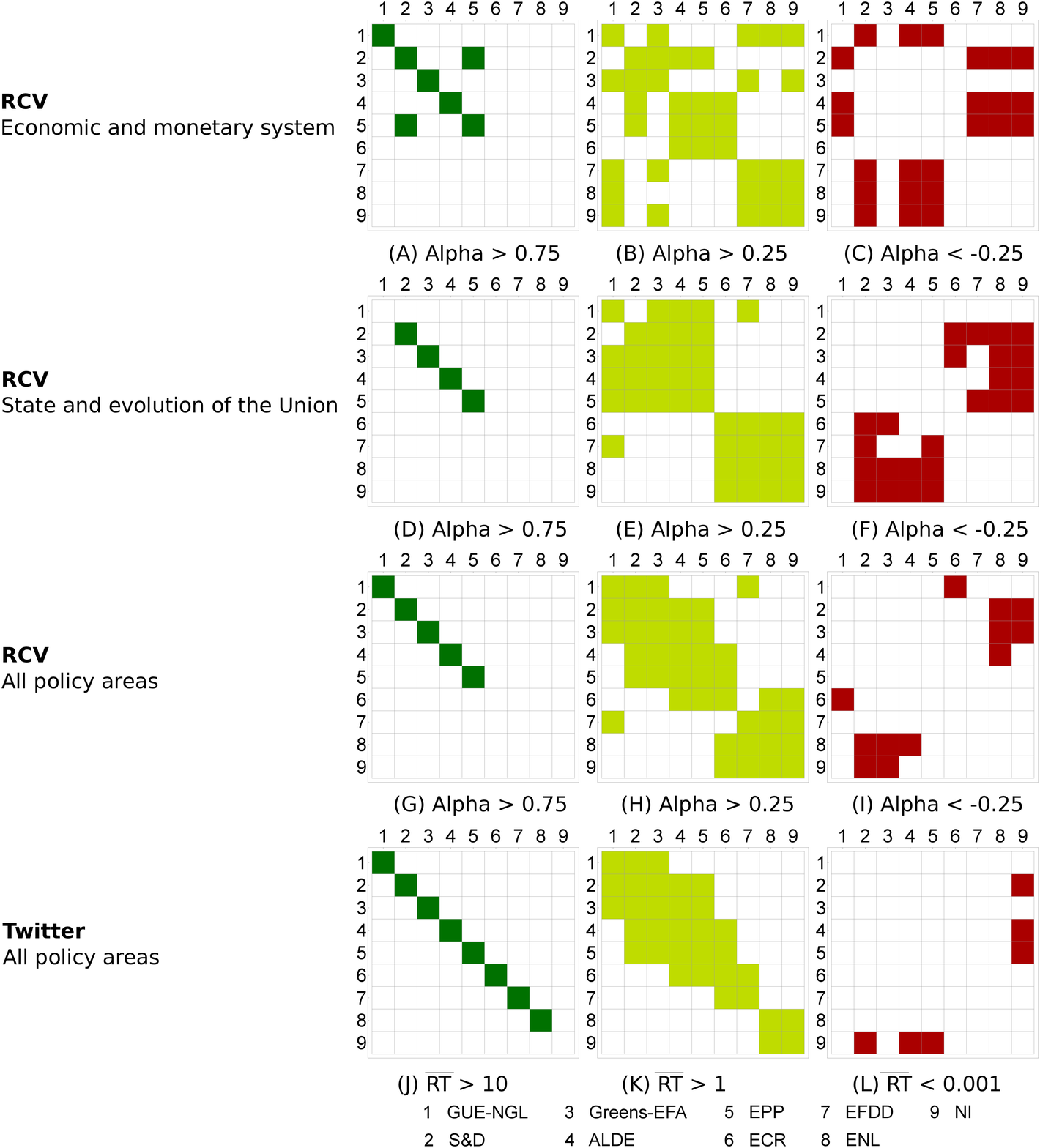}
\caption{{\bf Formation of coalitions ($\kalfa > 0$) and oppositions ($\kalfa < 0$)
in terms of the co-voting agreement (RCV).}
In the case of Twitter (the bottom charts),
coalitions are indicated by higher average retweets
($\rtavg$), and oppositions by lower average retweets.
}
\label{fig:rcv_twitter_blocks}
\end{figure}

The strongest cooperations in the European Parliament over all policy areas are shown in
Fig~\ref{fig:rcv_twitter_blocks}G. It comes as no surprise that the strongest
cooperations are within the groups (in the diagonal). Moreover, we again observe
\textit{GUE-NGL}, \textit{S\&D}, \textit{Greens-EFA}, \textit{ALDE}, and \textit{EPP} 
as the most cohesive groups. In Fig~\ref{fig:rcv_twitter_blocks}H, 
we observe coalitions forming along the diagonal, which represents the seating 
order in the European Parliament. Within this pattern, 
we observe four blocks of coalitions: 
on the left, between \textit{GUE-NGL}, \textit{S\&D}, and \textit{Greens-EFA}; 
in the center, between \textit{S\&D}, \textit{Greens-EFA}, \textit{ALDE},
and \textit{EPP};
on the right-center between \textit{ALDE}, \textit{EPP}, and \textit{ECR}; and finally, 
on the far-right between \textit{ECR}, \textit{EFDD}, \textit{ENL}, and \textit{NI}.
Fig~\ref{fig:rcv_twitter_blocks}I shows the strongest opposition between groups that
systematically disagree in voting. The strongest disagreements are between left- and
right-aligned groups, but not between the left-most and right-most groups,
in particular, between \textit{GUE-NGL} and \textit{ECR}, but also between \textit{S\&D} 
and \textit{Greens-EFA} on one side, and \textit{ENL} and \textit{NI} on the other.

In the area of \textit{Economic and monetary system} we see a strong cooperation 
between \textit{EPP} and \textit{S\&D} (Fig~\ref{fig:rcv_twitter_blocks}A), 
which is on a par with the cohesion of the most cohesive groups (\textit{GUE-NGL},
\textit{S\&D}, \textit{Greens-EFA}, \textit{ALDE}, and \textit{EPP}), 
and is above the cohesion of the other groups. As pointed out in the section
``\nameref{sec:coalitionpolicy}'', there is a strong separation in two blocks between
supporters and opponents of European integration, which is even more clearly observed 
in Fig~\ref{fig:rcv_twitter_blocks}B. On one hand, we observe cooperation between
\textit{S\&D}, \textit{ALDE}, \textit{EPP}, and \textit{ECR}, and on the other, 
cooperation between \textit{GUE-NGL}, \textit{Greens-EFA}, \textit{EFDD}, \textit{ENL}, 
and \textit{NI}. This division in blocks is seen again in
Fig~\ref{fig:rcv_twitter_blocks}C, which shows the strongest disagreements. 
Here, we observe two blocks composed of \textit{S\&D}, \textit{EPP}, and \textit{ALDE} 
on one hand, and \textit{GUE-NGL}, \textit{EFDD}, \textit{ENL}, and \textit{NI} 
on the other, which are in strong opposition to each other.

In the area of \textit{State and Evolution of the Union} we again observe a 
strong division in two blocks
(see Fig~\ref{fig:rcv_twitter_blocks}E). 
This is different to
the  \textit{Economic and monetary system}, however, where we observe a far-left 
and far-right cooperation, where the division is along the traditional left-right axis.

The patterns of coalitions forming on Twitter closely resemble those in 
the European Parliament. 
In Fig~\ref{fig:rcv_twitter_blocks}J we see that the strongest degrees of 
cooperation on Twitter are within the groups. The only group with low cohesion 
is the \textit{NI}, whose members have only seven retweets between them. 
The coalitions on Twitter follow the seating order in the 
European Parliament remarkably well
(see Fig~\ref{fig:rcv_twitter_blocks}K).
What is striking is that the same blocks 
form on the left, center, and on the center-right, both in 
the European Parliament and on Twitter. The largest
difference between the coalitions in 
the European Parliament and on Twitter is on 
the far-right, where we observe \textit{ENL} and \textit{NI} as isolated blocks.

\subsubsection{Relation between co-voting and retweeting}

\begin{figure}[!h]
\centering
\includegraphics[width=13cm]{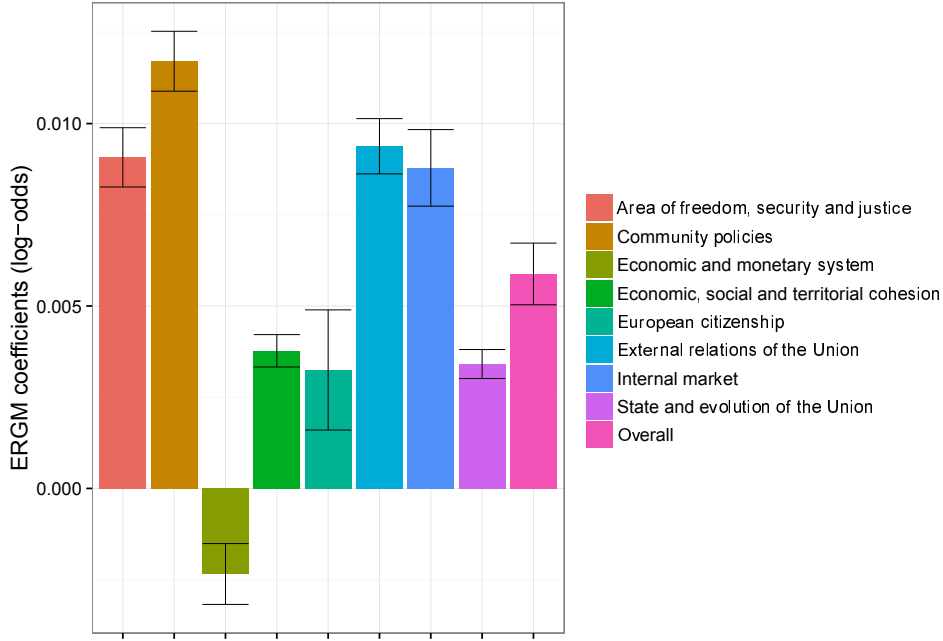}
\caption{{\bf Influence of retweeting on the co-voting of MEPs
as computed by ERGM.}
The influence is mostly positive, except for one policy area
(\textit{Economic and monetary system}), but very low.
The ERGM coefficients of $0.01$ correspond to an increase of probability
from the chance of $0.5$ to $0.503$.
}
\label{fig:twitter_ergm}
\end{figure}

The results shown in Fig~\ref{fig:twitter_ergm} quantify the extent to 
which communication in one social context (Twitter) 
can explain cooperation in another social context (co-voting in the European Parliament).
A positive value indicates that the matching behavior in the retweet network 
is similar to the one in the co-voting network,
specific for an individual policy area. On the other hand,
a negative value implies a negative ``correlation'' between the retweeting and
co-voting of MEPs in the two different contexts. 

The bars in Fig~\ref{fig:twitter_ergm} correspond to the coefficients from 
the edge covariate terms of the ERGM, describing the relationship between 
the retweeting and co-voting behavior of MEPs. The coefficients are aggregated 
for individual policy areas by means of a meta-analysis. 

Overall, we observe a positive correlation between retweeting and co-voting, 
which is significantly different from zero. The strongest positive correlations 
are in the areas \textit{Area of freedom, security and justice},
\textit{External relations of the Union}, and \textit{Internal markets}.
Weaker, but still positive, correlations are observed in the areas 
\textit{Economic, social and territorial cohesion}, \textit{European citizenship},
and \textit{State and evolution of the Union}. The only exception, with a 
significantly negative coefficient, is the area \textit{Economic and monetary system}.
This implies that in the area \textit{Economic and monetary system} 
we observe a significant deviation from the usual co-voting patterns. 
Results from section ``\nameref{sec:coalitionpolicy}'', 
confirm that this is indeed the case. 
Especially noteworthy are the coalitions 
between \textit{GUE-NGL} and \textit{Greens-EFA} on the left wing, 
and \textit{EFDD} and \textit{ENL} on the right wing.
In the section ``\nameref{sec:coalitionpolicy}'' we interpret these results 
as a combination of Euroscepticism on both sides, 
motivated on the left by a skeptical attitude towards the market orientation 
of the EU, and on the right by a reluctance to give up national sovereignty.

\section{Discussion}

We study cohesion and coalitions in the Eighth European Parliament by analyzing, 
on one hand, MEPs' co-voting tendencies and, on the other, their retweeting behavior. 

We reveal that the most cohesive political group in the European Parliament, 
when it comes to co-voting, is \textit{Greens-EFA}, closely followed by 
\textit{S\&D} and \textit{EPP}. This is consistent with what VoteWatch 
\cite{Votewatch2011seventh} reported for the Seventh European Parliament. 
The non-aligned 
(\textit{NI}) come out as the least cohesive group, followed by the Eurosceptic
\textit{EFDD}. Hix and Noury \cite{hix2009after} also report that nationalists and
Eurosceptics form the least cohesive groups in the Sixth European Parliament. 
We reaffirm most of these
results with both of the two employed methodologies. The only point where the two
methodologies disagree is in the level of cohesion for the left-wing 
\textit{GUE-NGL}, which is portrayed by ERGM as a much less cohesive group,
due to their relatively lower attendance rate.
The level of cohesion of the political groups is quite stable across different 
policy areas and similar conclusions apply.

On Twitter we can see results that are consistent with the RCV results for the
left-to-center political spectrum. The exception, which clearly stands out, is the
right-wing groups \textit{ENL} and \textit{EFDD} that seem to be the most cohesive ones.
This is the direct opposite of what was observed in the RCV data. 
We speculate that this phenomenon can be attributed to the fact that European right-wing groups,
on a European but also on a national level, rely to a large degree on social
media to spread their narratives critical of European integration.
We observed the same phenomenon recently during the Brexit campaign \cite{Grcar2016hirsch}. 
Along our interpretation the Brexit was ``won'' to some extent due to 
these social media activities, which are practically
non-existent among the pro-EU political groups. The fact that \textit{ENL} 
and \textit{EFDD} are the least cohesive groups in the European Parliament can be attributed to 
their political focus. It seems more important for the group to agree on its 
anti-EU stance and to call for independence and sovereignty, and much less important 
to agree on other issues put forward in the parliament. 

The basic pattern of coalition formation, with respect to co-voting, can already be seen 
in Fig~\ref{fig:covoting_retweeting_network}A: the force-based layout almost completely
corresponds to the seating order in the European Parliament 
(from the left- to the right-wing groups). 
A more thorough examination shows that the strongest cooperation can be observed, 
for both methodologies, between \textit{EPP}, \textit{S\&D}, and \textit{ALDE}, 
where \textit{EPP} and \textit{S\&D} are the two largest groups, while the liberal
\textit{ALDE} plays the role of an intermediary in this context. 
On the other hand, the role of an intermediary between the far-left \textit{GUE-NGL} 
and its center-left neighbor, \textit{S\&D}, is played by the 
\textit{Greens-EFA}.
These three parties also form a strong coalition in the European Parliament. 
On the far right of the spectrum, the non-aligned, \textit{EFDD}, and \textit{ENL} 
form another coalition. This behavior was also observed by Hix et al.
\cite{hix2007democratic}, stating that alignments on the left and right have in recent 
years replaced the ``Grand Coalition'' between the two large blocks of 
Christian Conservatives (\textit{EPP}) and Social Democrats (\textit{S\&D}) 
as the dominant form of finding majorities in the European Parliament. 
When looking at the policy 
area \textit{Economic and monetary system}, we see the same coalitions. However,
interestingly, \textit{EFDD}, \textit{ENL}, and \textit{NI} often co-vote with 
the far-left \textit{GUE-NGL}. This can be attributed to a certain degree of
Euroscepticism on both sides: as a criticism of capitalism, on one hand, and as 
the main political agenda, on the other. This pattern was also discovered by Hix et al.
\cite{hix2007democratic}, who argued that these coalitions emerge from a form of
government-opposition dynamics, rooted at the national level, but also reflected 
at the European level. 

When studying coalitions on Twitter, the strongest coalitions can be observed on the 
right of the spectrum (between \textit{EFDD}, \textit{ECR}, \textit{ENL}, 
and \textit{NI}). This is, yet again, in contrast to what was observed in the RCV data. 
The reason lies in the anti-EU messages they tend to collectively spread (retweet) 
across the network. This behavior forms strong retweet ties, not only within, but also
between, these groups. For example, MEPs of \textit{EFDD} mainly retweet MEPs from
\textit{ECR} (with the exception of MEPs from their own group). In contrast to these 
right-wing coalitions, we find the other coalitions to be consistent with what is 
observed in the RCV data. The strongest coalitions on the left-to-center part of 
the axis are those between \textit{GUE-NGL}, \textit{Greens-EFA}, and \textit{S\&D}, 
and between \textit{S\&D}, \textit{ALDE}, and \textit{EPP}. These results reaffirm 
the role of \textit{Greens-EFA} and \textit{ALDE} as intermediaries, not only in 
the European Parliament
but also in the debates on social media.

Last, but not least, with the ERGM methodology we measure the extent to which the 
retweet network can explain the co-voting activities in the European Parliament. 
We compute this for 
each policy area separately and also over all RCVs. We conclude that the retweet 
network indeed matches the co-voting behavior, with the exception of one specific 
policy area. In the area \textit{Economic and monetary system}, the links in the 
(overall) retweet network do not match the links in the co-voting network. Moreover, 
the negative coefficients imply a radically different formation of coalitions 
in the European Parliament. This is consistent with the results 
in Figs~\ref{fig:coalition_overall_ka} 
and \ref{fig:coalition_overall_ergm}
(the left-hand panels), and is also observed in
Fig~\ref{fig:rcv_twitter_blocks} (the top charts). 
From these figures we see that in this particular case,
the coalitions are also formed  between the right-wing groups and the far-left GUE-NGL. 
As already explained, we attribute this to the degree of Euroscepticism that these groups
share on this particular policy issue.

\section{Conclusions}

In this paper we analyze (co-)voting patterns and social behavior
of members of the European Parliament, as well as the interaction
between these two systems. More precisely, we analyze a set of 2535
roll-call votes as well as the tweets and retweets of members of the MEPs
in the period from October 2014 to February
2016. The results indicate a considerable level of correlation between
these two complex systems. This is consistent with previous findings
of Cherepnalkoski et al. \cite{Cherepnalkoski2016retweet}, who
reconstructed the adherence of MEPs to their respective political or
national group solely from their retweeting behavior.

We employ two different methodologies to quantify the co-voting
patterns: Krippendorff's \alfa\, and ERGM. They were developed in
different fields of research, use different techniques, and are based
on different assumptions, but in general they yield consistent
results. However, there are some differences which have consequences
for the interpretation of the results.

\alfa\, is a measure of agreement, designed as a generalization
of several specialized measures, that can compare different numbers of
observations, in our case roll-call votes. It only considers
yes/no votes. Absence and abstention by MEPs is ignored. Its baseline
($\kalfa=0$), i.e., co-voting by chance, is computed from the yes/no
votes of all MEPs on all RCVs.

ERGMs are used in social-network analyses to determine factors
influencing the edge formation process. In our case an edge between
two MEPs is formed when they cast the same yes/no vote within
a RCV. It is assumed that a priori each MEP can form a link
with any other MEP. No assumptions about the presence or absence of
individual MEPs in a voting session are made. Each RCV is analyzed as
a separate binary network. The node set is thereby kept constant for
each RCV network. While the ERGM departs from the originally observed
network, where MEPs who didn't vote or abstained appear as isolated
nodes, links between these nodes are possible within the network
sampling process which is part of the ERGM optimization process. The
results of several RCVs are aggregated by means of the meta-analysis
approach. The baseline (ERGM coefficients $=0$), i.e., co-voting by
chance, is computed from a large sample of randomly generated
networks.

These two different baselines have to be taken into account when
interpreting the results of \alfa\, and ERGM.  In a typical voting
session, 25\% of the MEPs are missing or abstaining. When assessing
cohesion of political groups, all \alfa\, values are well above the
baseline, and the average $\kalfa=0.7$. The average ERGM cohesion
coefficients, on the other hand, are around the baseline. The
difference is even more pronounced for groups with higher
non-attendance/abstention rates like \textit{GUE-NGL} (34\%) and
\textit{NI} (40\%). When assessing strength of coalitions between
pairs of groups, \alfa\, values are balanced around the baseline,
while the ERGM coefficients are mostly negative.  The ordering of
coalitions from the strongest to the weakest is therefor different
when groups with high non-attendance/abstention rates are involved.

The choice of the methodology to asses cohesion and coalitions is
not obvious. Roll-call voting is used for decisions which demand a
simple majority only.
One might however argue that non-attendance/abstention corresponds
to a no vote, or that absence is used strategically.
Also, the importance of individual votes, i.e., how high on the
agenda of a political group is the subject, affects their attendance,
and consequently the perception of their cohesion and the potential to
act as a reliable coalition partner.

\section*{Acknowledgments}
This work was supported in part by the EC projects SIMPOL (no. 610704)
and DOLFINS (no. 640772), 
and by the Slovenian ARRS programme Knowledge Technologies (no. P2-103).



\end{document}